\documentclass[conference,a4paper]{IEEEtran}
\IEEEoverridecommandlockouts

\usepackage[hidelinks]{hyperref}
\usepackage[cmex10]{amsmath}
\usepackage{amssymb,amsfonts}
\usepackage{dblfloatfix}

\usepackage[ruled,vlined]{algorithm2e}
\usepackage{graphicx}
\graphicspath{{Figuress/PDF/}{Figures/PNG/}}

\usepackage{booktabs}
\usepackage{siunitx}
\usepackage[numbers,compress]{natbib}
\usepackage{texnames}
\usepackage{bm,bbm}
\usepackage{orcidlink}
\usepackage{multirow, makecell}
\usepackage{subfigure}

\begin{document}
\title{\uppercase{{Fourier-Modulated Implicit Neural Representation for Multispectral Satellite Image Compression}}

}

\author{\IEEEauthorblockN{Woojin Cho\textsuperscript{*}, \;\;\; Steve Andreas Immanuel\textsuperscript{*}, \;\;\; Junhyuk Heo, \;\;\; Darongsae Kwon}
\IEEEauthorblockA{
TelePIX \\
07330, Seoul, South Korea
\\
\{woojin, steve, hjh1037, darong.kwon\}@telepix.net \\
\thanks{\textsuperscript{*}These authors contributed equally.}
}

}

\maketitle
\begin{abstract}
Multispectral satellite images play a vital role in agriculture, fisheries, and environmental monitoring. However, their high dimensionality, large data volumes, and diverse spatial resolutions across multiple channels pose significant challenges for data compression and analysis. This paper presents \mbox{ImpliSat}, a unified framework specifically designed to address these challenges through efficient compression and reconstruction of multispectral satellite data. ImpliSat leverages Implicit Neural Representations (INR) to model satellite images as continuous functions over coordinate space, capturing fine spatial details across varying spatial resolutions. Furthermore, we introduce a Fourier modulation algorithm that dynamically adjusts to the spectral and spatial characteristics of each band, ensuring optimal compression while preserving critical image details.
\end{abstract}


\section{Introduction}
\label{intro}

Satellite data is essential for research and applications such as climate change~\citep{yang2013role, norris2016evidence} and marine ecosystem monitoring~\cite{masek2001stability}. In particular, multispectral satellite imagery (MSI) enables a detailed analysis of soil conditions~\cite{hassan2015assessment}, vegetation distribution~\cite{dwyer2000global, ma2008detecting}, and natural resource management~\cite{pettorelli2019satellite}, as it contains information collected in various wavelength bands.

In order to collect MSI, satellites are deployed to orbit Earth at a certain altitude and operate via ground stations. As the satellite orbits, it gathers data from Earth's surface using broad ranges of sensor and stores them in its onboard computer storage. This data is then transmitted back to the ground station for further processing. However, communication between the satellite and the ground station is limited to the orbital pass of the satellite, \textit{i.e.} periods when the satellite is within the line of sight of the ground station. Consequently, satellites must limit data collection to ensure it can be fully transmitted within the available communication window. Any excess data may need to be discarded to free up storage for new observations or held for transmission during a subsequent orbital pass. This motivates us to develop a specific compression algorithm for remote sensing image data.

\begin{figure}[t]
\centering
\includegraphics[width=1\columnwidth]{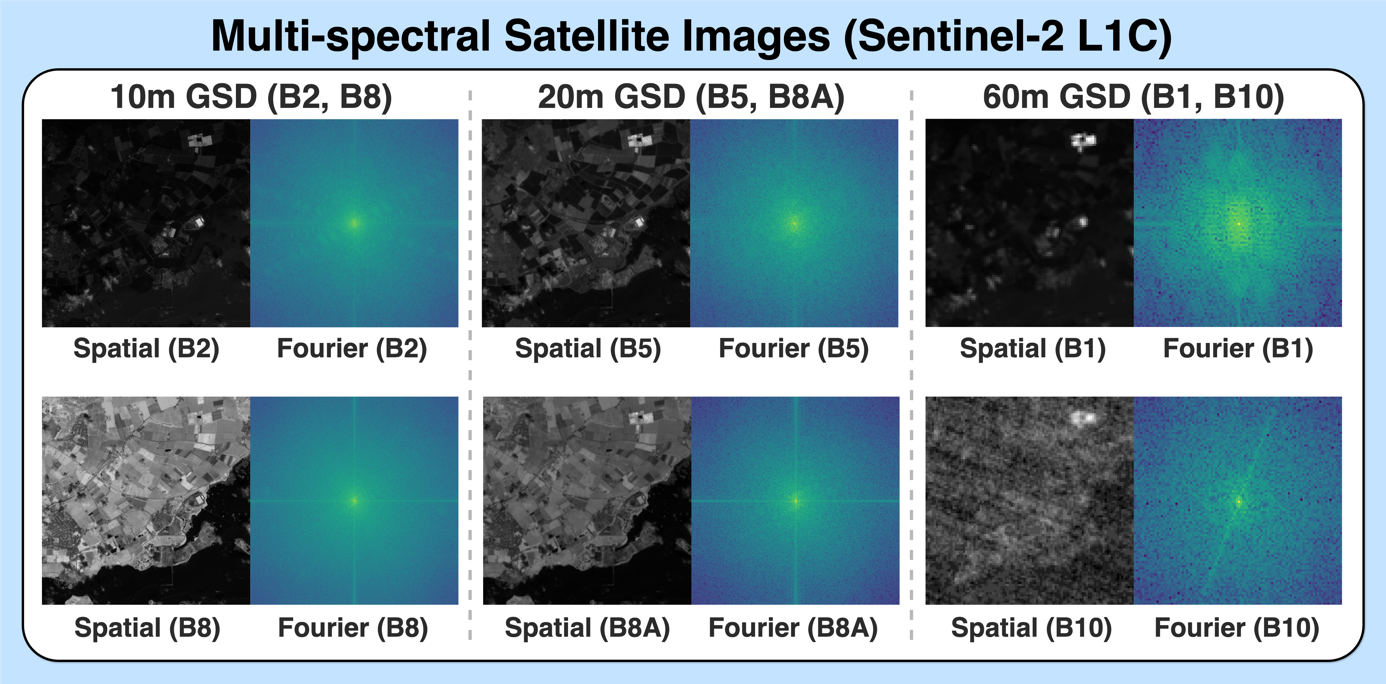}
\caption{Visualization of Sentinel-2 L1C MSI (London) in spatial and frequency domains. The images include bands at 10m GSD (B2, B8), 20m GSD (B5, B8A), and 60m GSD (B1, B10), illustrating the varying levels of detail captured at different resolutions.}
\label{fig:msi_london}
\end{figure}

MSI has multiple bands, and each band can have a different ground sample distance (GSD). Consequently, the frequency spectrum varies significantly as shown in Fig. \ref{fig:msi_london}. These differences in spatial resolution and spectral characteristics present challenges for efficient data representation and compression.

We choose to use Implicit Neural Representation (INR) to develop the compression network since it can render continuous spatial data at arbitrary resolutions by representing the data as coordinate-based functions. This property enables efficient compression of high-resolution imagery, making it particularly effective for reconstructing data at varying scales. However, existing INR methods have primarily focused on RGB images~\cite{sitzmann2020implicit, strumpler2022inrcompress, mudiyanselage2025unveiling}, 3D modeling~\cite{mildenhall2020nerf, pumarola2021d} and signal representation~\cite{cho2025neural} with limited exploration of multispectral satellite data. Unlike RGB or 3D data, MSI is characterized by varying wavelength bands and differing spatial resolutions across channels, presenting unique challenges.

\begin{figure*}[ht]
\centering
\includegraphics[width=1.8\columnwidth]{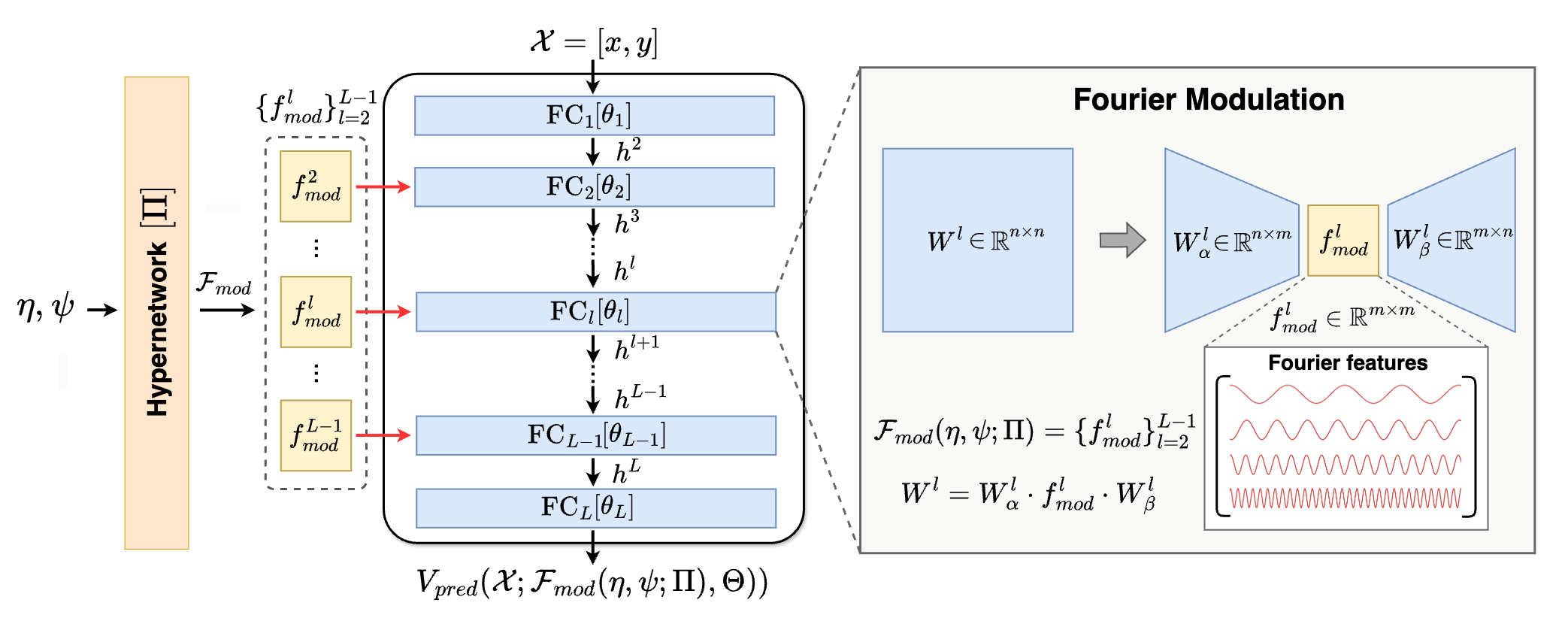}
\caption{Overall architecture of ImpliSat. The left part shows the INR backbone, which takes spatial coordinates as input and is conditioned on resolution $(\eta)$ and channel information $(\psi)$. The right part demonstrates the Fourier modulation process, which effectively represents the MSI data.}
\label{fig:archi}
\end{figure*}

Existing INR approaches typically apply fixed Fourier features uniformly across all bands to first map the coordinates into higher-dimensional space. However, the frequency spectrum difference in channels with 10m, 20m, and 60m GSD suggests that it may be beneficial to adapt different Fourier features depending on the GSD. This approach enables a more accurate and efficient representation of the data.

In this paper, we introduce \textbf{Impli}cit Neural Representations for Multispectral \textbf{Sat}ellite Images (\textbf{ImpliSat}), a novel INR-based compression framework specifically designed for MSI. ImpliSat models MSI data as coordinate-based functions, compressing and reconstructing the data while accounting for the differences in spatial resolution and channel information.
Additionally, our ImpliSat framework introduces a hypernetwork-based Fourier modulation technique that dynamically generates appropriate Fourier bases tailored to each spectral band resolution. By conditioning the modulation on the unique resolution and channel information of each band, our approach ensures that each spectral channel is represented optimally, enhancing the efficiency of compression and the accuracy of reconstruction.

\section{Preliminary: Multispectral Satellite Images and Compression}
Unlike standard RGB images, MSI encompasses multiple spectral bands, often exceeding ten. Each spectral band has different pixel value range and covers a specific range of wavelengths with different GSD depending on the sensor (cf. Fig.~\ref{fig:msi_london}). 
GSD denotes the actual ground distance between adjacent pixels (10m GSD means 1 pixel in the image represents 10m in the actual distance). 
The bigger the GSD, the lower the spatial resolution, which means there will be fewer details visible in the image.

Traditional image compression algorithms, such as JPEG~\cite{wallace1992jpeg} and PNG~\cite{boutell1997png}, are optimized for uniform resolution images and do not accommodate the complex, multi-resolution structure of MSI. More importantly, they are designed specifically for images with 3 (RGB) and 4 (RGBA) channels (for JPEG and PNG, respectively), thus cannot be applied for MSI.
Existing deep learning-based image compression techniques~\cite{dupont2022coin++} are also not designed to handle the diverse spatial resolutions and pixel value ranges inherent in MSI, leading to limitations in their performance (we provide further details in Section~\ref{sec:experiments}).

\section{Proposed Methods}\label{sec:proposed_method}
ImpliSat is built based on modulated INR frameworks. Our key contribution is Fourier modulation, a novel modulation technique that leverages spectral information and low-rank adaptation~\cite{hu2021lora, cho2024hypernetwork, cho2025fastlrnr} to efficiently handle the multi-band and multi-resolution nature of MSI data.
The overall architecture is shown in Fig. \ref{fig:archi}.
ImpliSat consists of two main components: i) Hypernetwork that generates Fourier modulations based on resolution and channel information, and ii) SIREN-based backbone INR model~\cite{sitzmann2020implicit} that uses these modulation vectors to represent each band of the multispectral image. 

\subsection{INR Networks with Sinusoidal Activation}
The main INR network of ImpliSat is a multilayer perceptron (MLP) with $L$ layers parameterized by $\Theta = \{\theta_l\}_{l=1}^L$, which takes spatial coordinates $\mathcal{X} \in \mathbb{R}^2$ as input and predicts the corresponding pixel value $V_{gt}(\mathcal{X}, \eta, \psi)$.
Each layer of the INR backbone consists of fully connected layers, where the $l$-th layer is parameterized by $\theta_l = \{W^l, b^l\}$, with $W^l \in \mathbb{R}^{n \times n}$ as the weight matrix and $b^l \in \mathbb{R}^{n}$ as the bias term. The hidden state at $(l+1)$-th layer is computed using sinusoidal activations \cite{sitzmann2020implicit} as $h^{l+1} = \sin(W^l \cdot h^l + b^l)$, where $h^l$ is the hidden state at $l$-th layer. This sinusoidal activation allows the model to learn high-frequency information, which is crucial for representing complex features in MSI data. The whole network is trained end-to-end to minimize the L2 distance between the ground truth pixel values of the MSI $V_{gt}$ and the predicted reconstruction values $V_{pred}$.

\subsection{Hypernetworks and Fourier Modulations}
The hypernetwork parameterized by $\Pi$ in ImpliSat generates Fourier modulations, following the approach introduced in~\cite{shi2024improved}, to adapt the INR model to varying resolutions $\eta$ and channels $\psi$. Specifically, the hypernetwork takes the following inputs, $\eta$ and $\psi$\footnote{The values for $\eta$ and $\psi$ follow Sentinel-2 data}:
\begin{itemize}
\item \textbf{Resolution information} $(\eta)$: Represents the spatial resolution (GSD) of the input MSI, which can take values $\eta \in \{10, 20, 60\}$.
\item \textbf{Channel information} $(\psi)$: Represents one of the 13 different spectral channels in the MSI data, encoded as a one-hot vector.
\end{itemize}
For each hidden layer of the backbone INR $\Theta$, the hypernetwork first generates Fourier bases as follows:
\begin{align}
\label{eq:modulation}
\mathcal{F}_{mod}(\eta, \psi ; \Pi) = \{{\Omega}^l, \varphi^l \}_{l=2}^{L-1}, 
\end{align}
where ${\Omega}^l \in \mathbb{R}^{m \times m}$ and $\varphi^l \in \mathbb{R}^{m \times m}$ are the Fourier frequency matrix and phase matrix for the $l$-th layer, respectively. Then, we sample an $m$-dimensional vector from $\mathcal{U}(-2\pi, 2\pi)$ and stack it $m$ times to form matrix $\mathcal{Z} \in \mathbb{R}^{m \times m}$. Finally, the Fourier modulations can be calculated using a $cosine$ function as follows:
\begin{align}
\label{eq:Fourier_modulation}
\{f^l_{mod}\}_{l=2}^{L-1} = \{\cos({\Omega}^l \odot \mathcal{Z} + {\varphi}^l)\}_{l=2}^{L-1},
\end{align}
where $\odot$ denotes element-wise multiplication. As shown in Equation~\eqref{eq:Fourier_modulation}, this approach dynamically adapts to different resolutions and bands, ensuring effective representation of multispectral satellite images.

\subsection{INR Networks with Fourier-modulated Weights}
We apply the Fourier modulations to the weights of each hidden layer of the backbone INR using the following formula:
\begin{align}
 W^l = W_{\alpha}^l \cdot f_{mod}^l \cdot W_{\beta}^l,
\end{align}
where $W^l$ is decomposed into matrix multiplications of $W_{\alpha}^l \in \mathbb{R}^{n \times m}$, $f_{mod}^l \in \mathbb{R}^{m \times m}$, and $W_{\beta}^l \in \mathbb{R}^{m \times n}$. This serves two purposes: 1) It allows us to modulate the backbone INR using the Fourier modulations produced by the hypernetwork. This Fourier modulation technique allows each layer of the backbone INR model to represent the
data using Fourier bases adapted to the specific resolution and channel information of the target MSI. 2) It significantly reduces the computational cost by applying low-rank adaptation \cite{hu2021lora}. We specifically choose $m << n$, therefore, the only trainable parts of the backbone INR are only low-rank matrices $W_{\alpha}^l$ and $W_{\beta}^l$.



\section{Experiments}\label{sec:experiments}

In this section, we demonstrate the performance of the ImpliSat framework using diverse MSI data for compression and reconstruction tasks. Additionally, we explore the results of Fourier modulation across various resolutions.

\begin{table}[ht!]
\small
\centering
\caption{Geographic Coordinates of Benchmark Dataset}
\begin{tabular}{llllll}
\specialrule{1pt}{2pt}{2pt}
\textbf{Dataset} & \textbf{Latitude} & \textbf{Longitude} & \textbf{Environments} \\
\specialrule{1pt}{2pt}{2pt}
\textbf{Cairo} & 30$^{\circ}$01'29"N & 31$^{\circ}$55'18"E & Desert \\
\textbf{Merapi} & 07$^{\circ}$32'29"N & 110$^{\circ}$26'46"E & Volcano \\
\textbf{London} & 51$^{\circ}$26'11"N & 00$^{\circ}$34'55"E & Urban \\
\textbf{Seoul} & 37$^{\circ}$31'30"N & 126$^{\circ}$55'36"E & Urban \\
\textbf{Hawaii} & 19$^{\circ}$31'12"N & 154$^{\circ}$47'08"W & Marine \\
\specialrule{1pt}{2pt}{2pt}
\end{tabular}\label{tbl:dataset_coord}
\end{table}

\begin{figure}[ht!]
\centering
\renewcommand{\arraystretch}{0.5} 
\setlength{\tabcolsep}{1pt} 

\begin{tabular}{@{}c@{\hspace{1.5mm}} cccc@{}}
& GT & Shift & Scale & \textbf{Ours} \\[1.5mm]

\rotatebox{90}{\parbox[c]{2cm}{\centering B2 (10m)}} &
\includegraphics[width=0.23\columnwidth]{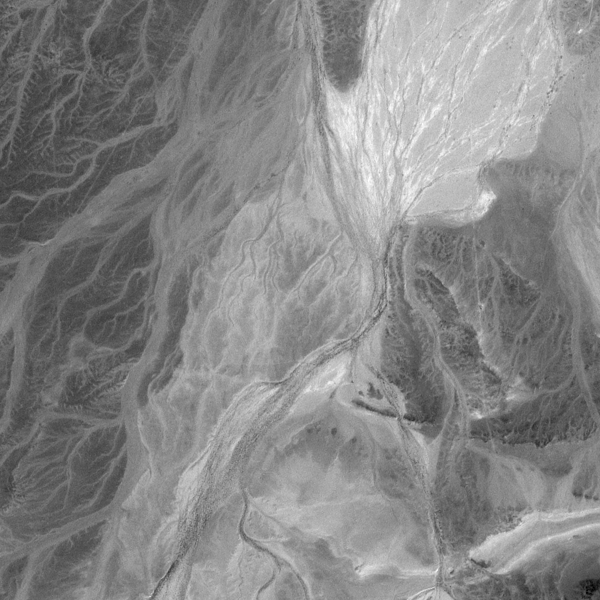} &
\includegraphics[width=0.23\columnwidth]{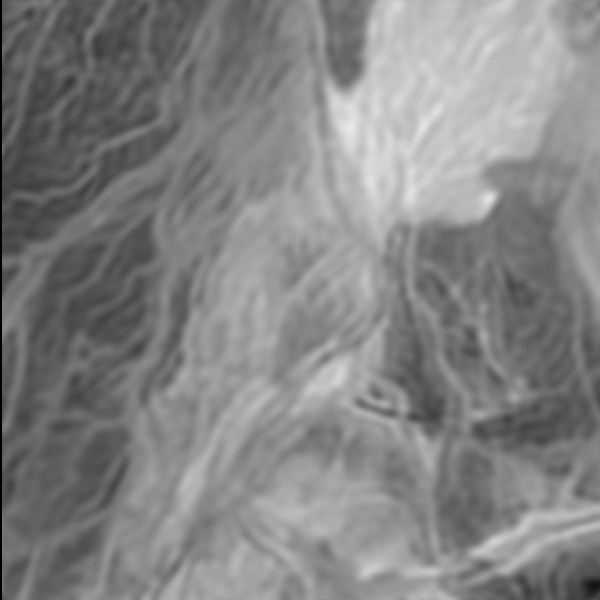} &
\includegraphics[width=0.23\columnwidth]{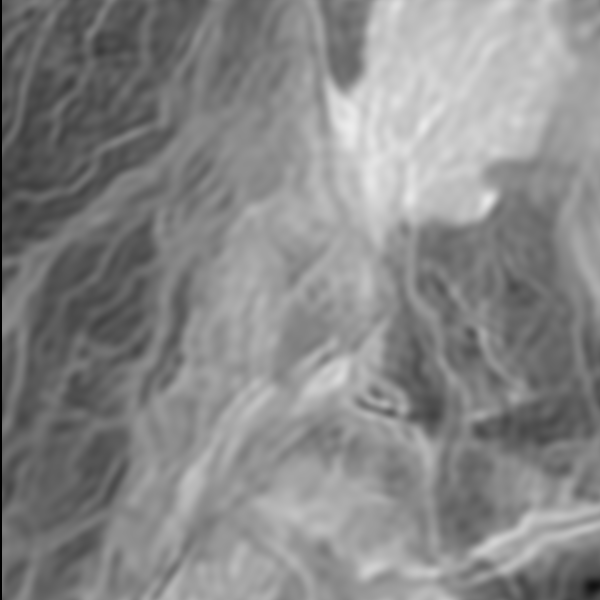} &
\includegraphics[width=0.23\columnwidth]{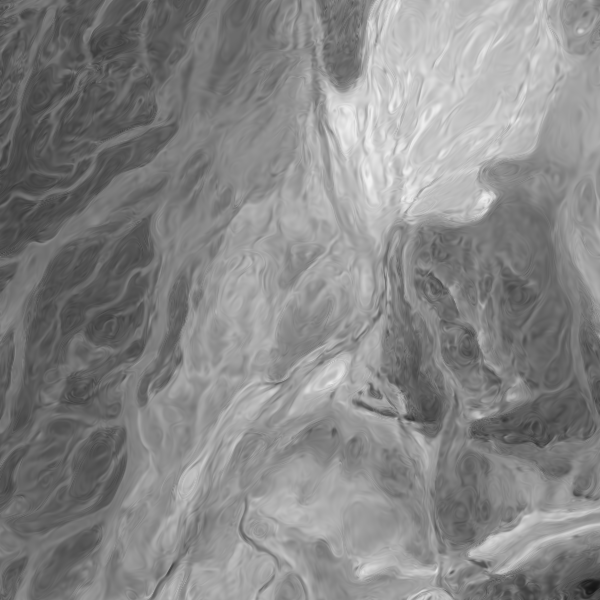} \\
&  & 29.257 & 28.777 & \textbf{32.124} \\[1mm]

\rotatebox{90}{\parbox[c]{2cm}{\centering B3 (10m)}} &
\includegraphics[width=0.23\columnwidth]{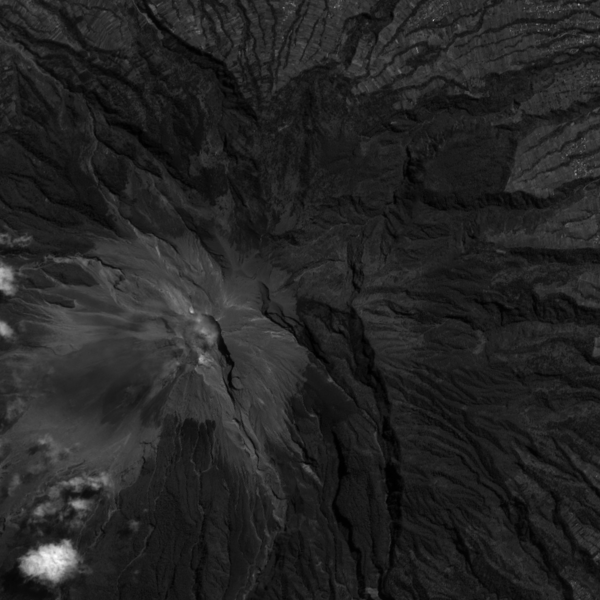} &
\includegraphics[width=0.23\columnwidth]{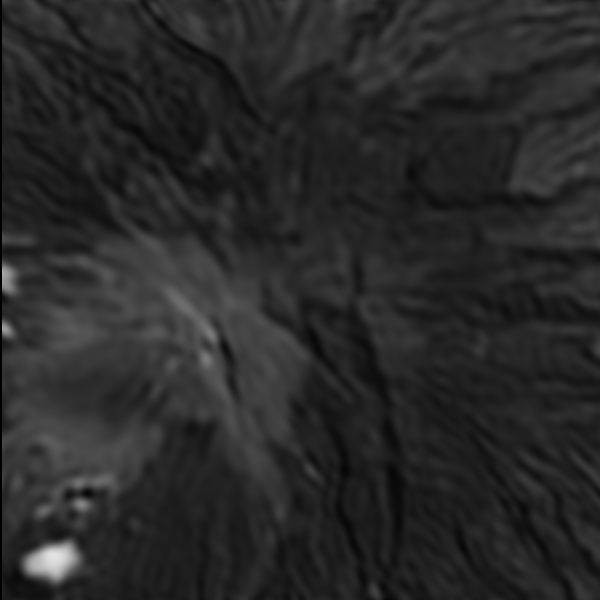} &
\includegraphics[width=0.23\columnwidth]{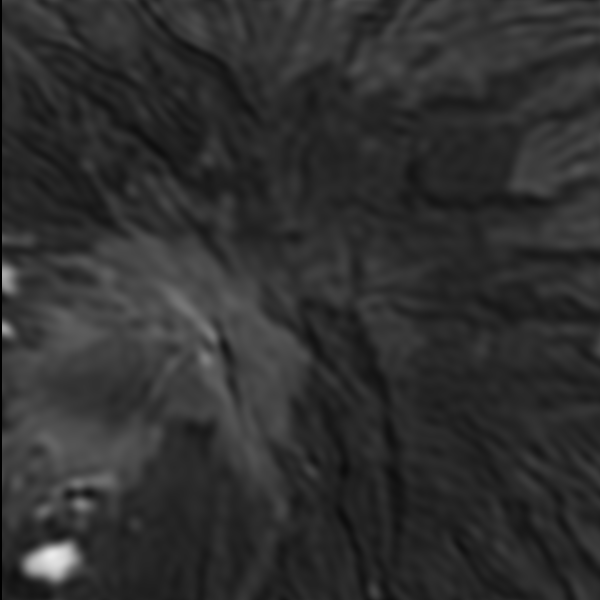} &
\includegraphics[width=0.23\columnwidth]{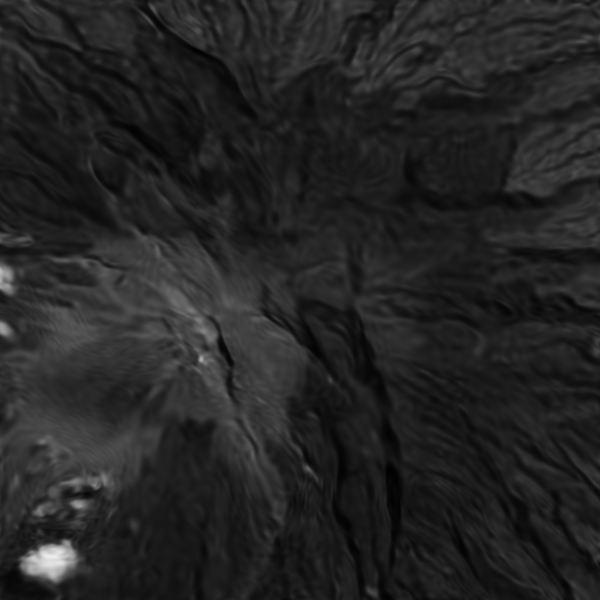} \\
&  & 31.401 & 31.059 & \textbf{33.398} \\[1mm]

\rotatebox{90}{\parbox[c]{2cm}{\centering B7 (20m)}} &
\includegraphics[width=0.23\columnwidth]{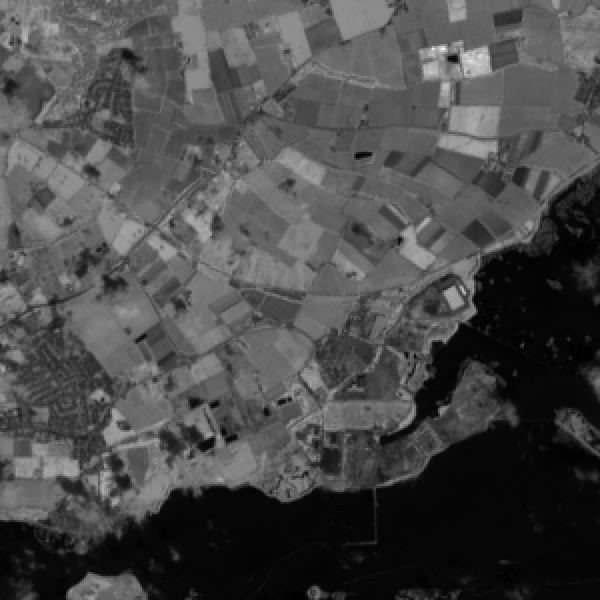} &
\includegraphics[width=0.23\columnwidth]{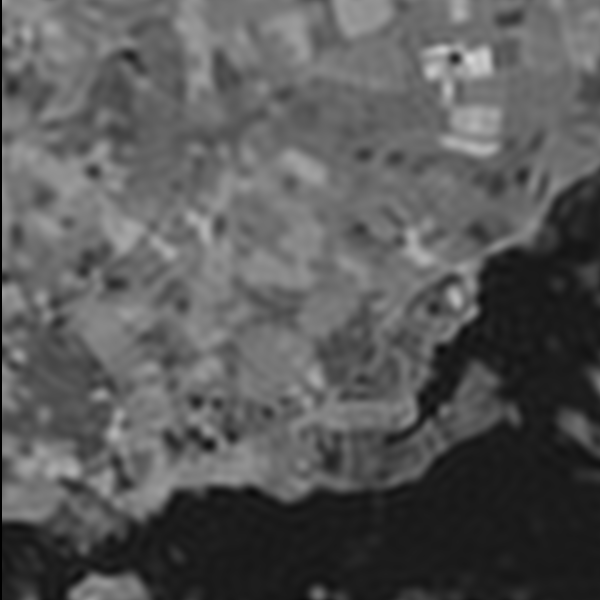} &
\includegraphics[width=0.23\columnwidth]{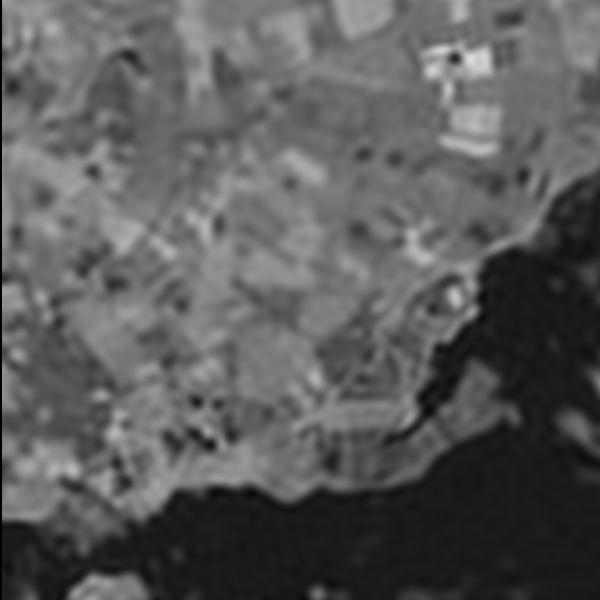} &
\includegraphics[width=0.23\columnwidth]{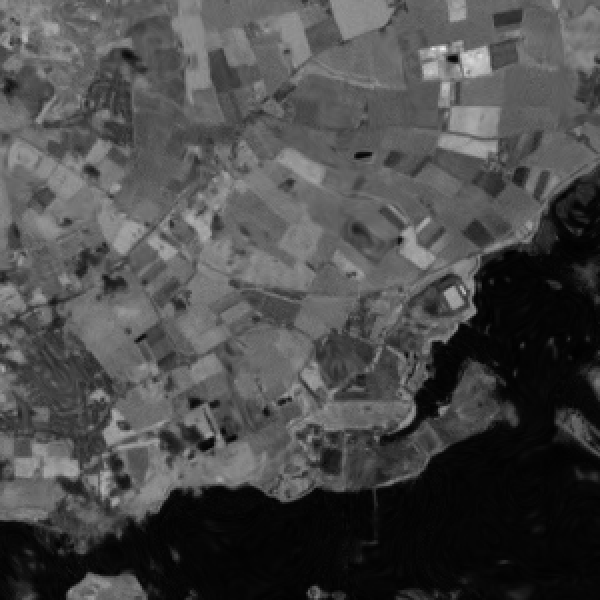} \\
&  & 28.712 & 28.532 & \textbf{35.751} \\[1mm]

\rotatebox{90}{\parbox[c]{2cm}{\centering B8A (20m)}} &
\includegraphics[width=0.23\columnwidth]{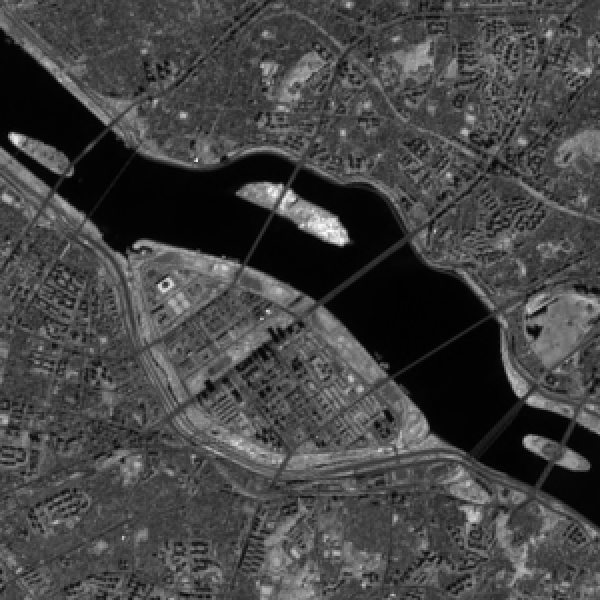} &
\includegraphics[width=0.23\columnwidth]{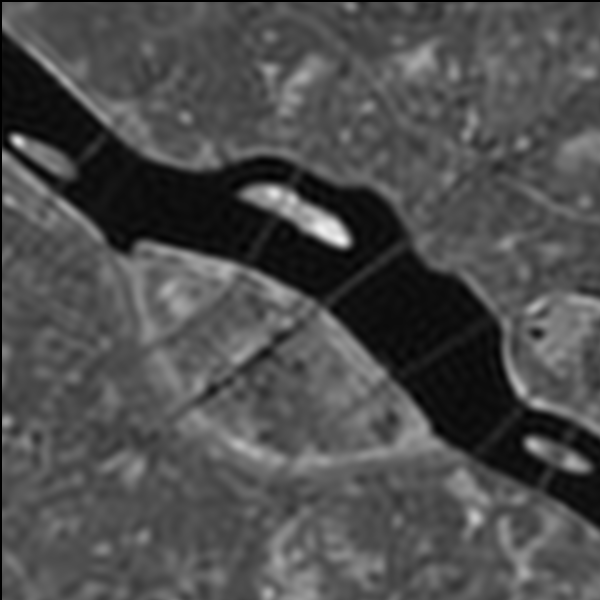} &
\includegraphics[width=0.23\columnwidth]{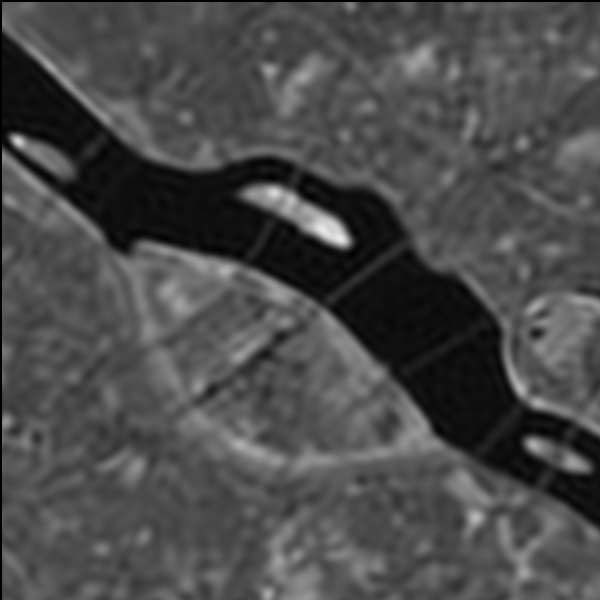} &
\includegraphics[width=0.23\columnwidth]{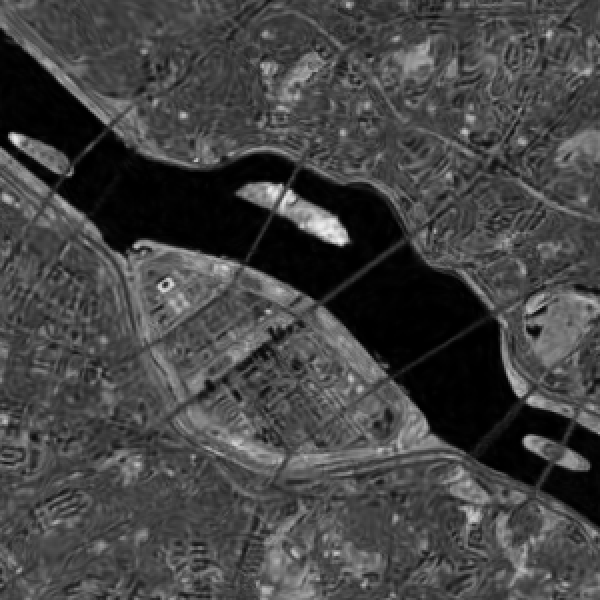} \\
&  & 24.400 & 24.195 & \textbf{29.994} \\[1mm]

\rotatebox{90}{\parbox[c]{2cm}{\centering B1 (60m)}} &
\includegraphics[width=0.23\columnwidth]{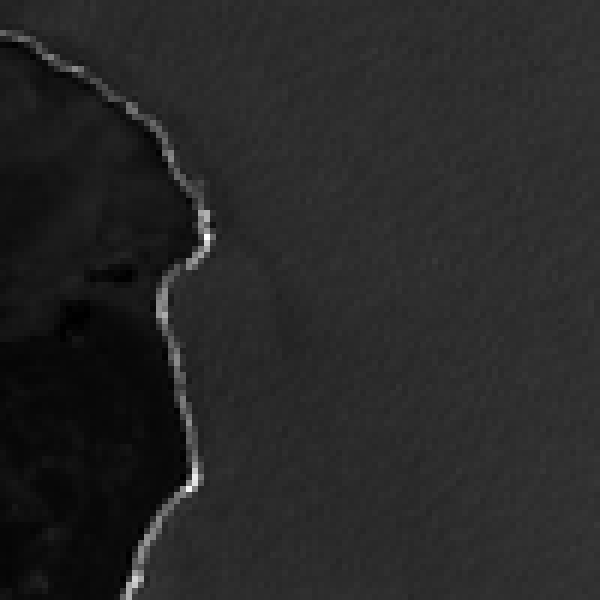} &
\includegraphics[width=0.23\columnwidth]{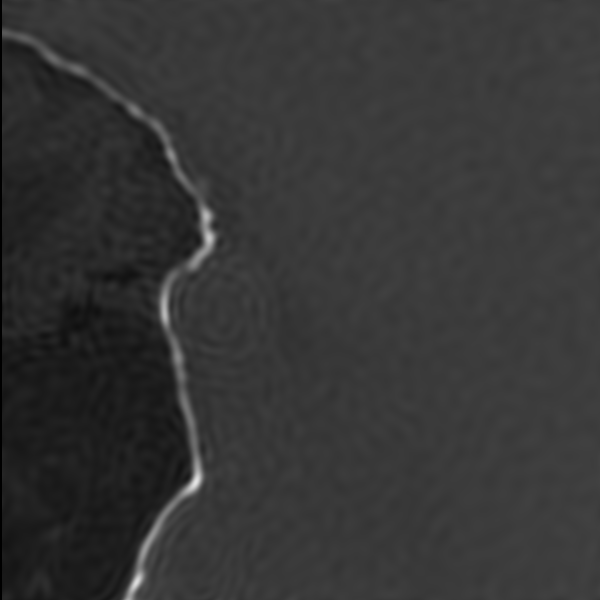} &
\includegraphics[width=0.23\columnwidth]{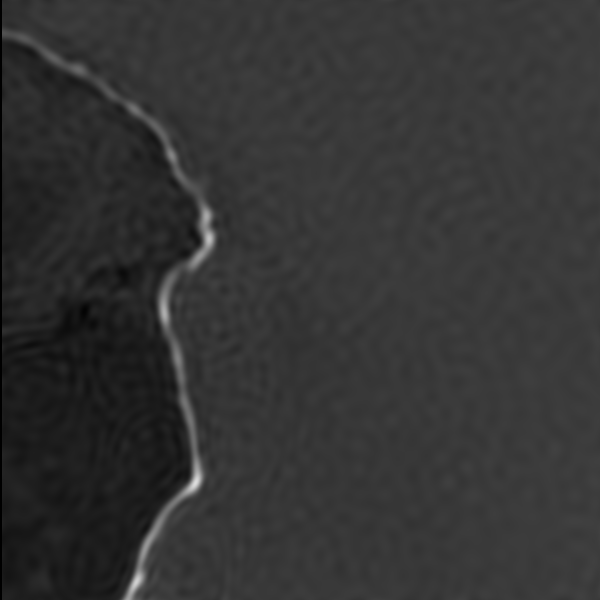} &
\includegraphics[width=0.23\columnwidth]{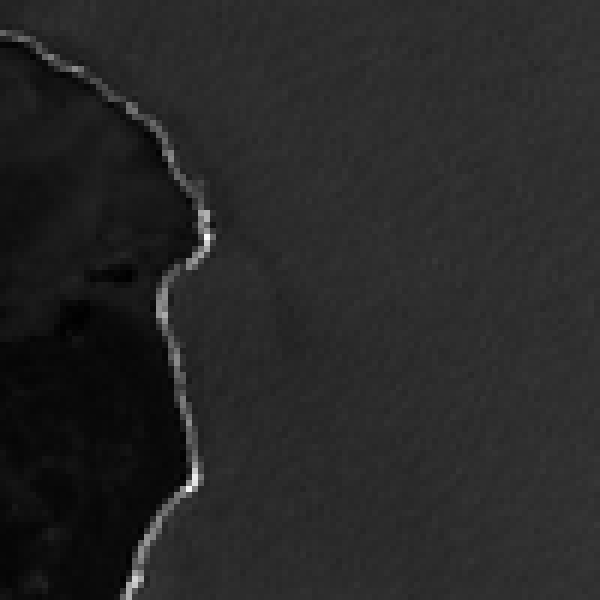} \\
&  & 33.754 & 33.438 & \textbf{48.434} \\[1mm]

\end{tabular}

\caption{Comparison of ground truth, shift modulation, scale modulation, and our proposed method with PSNR on all MSI.}
\label{fig:main_results}
\end{figure}

\begin{table*}[t]
\centering
\caption{PSNR$\uparrow$ and MSE$\downarrow$ comparison for the Fourier modulation method (ours) against baseline methods}
\small
\begin{tabular}{lllllllllll}
\specialrule{1pt}{2pt}{2pt}
\multirow{3}{*}{\textbf{Methods}} & \multicolumn{2}{c}{\textbf{London}} & \multicolumn{2}{c}{\textbf{Seoul}} & \multicolumn{2}{c}{\textbf{Merapi}} & \multicolumn{2}{c}{\textbf{Hawaii}} & \multicolumn{2}{c}{\textbf{Cairo}} \\ \cmidrule(lr){2-3} \cmidrule(lr){4-5} \cmidrule(lr){6-7} \cmidrule(lr){8-9} \cmidrule(lr){10-11}
 & \multicolumn{1}{c}{PSNR} & \multicolumn{1}{c}{MSE} & \multicolumn{1}{c}{PSNR} & \multicolumn{1}{c}{MSE} & \multicolumn{1}{c}{PSNR} & \multicolumn{1}{c}{MSE} & \multicolumn{1}{c}{PSNR} & \multicolumn{1}{c}{MSE} & \multicolumn{1}{c}{PSNR} & \multicolumn{1}{c}{MSE} \\
\specialrule{1pt}{2pt}{2pt}
\textbf{Shift} & 30.252 & 9.437e-4 & 28.115 & 1.543e-3 & 28.567 & 1.391e-3 & 29.418 & 1.143e-3 & 29.966 & 1.008e-3 \\ \cmidrule(lr){1-11}
\textbf{Scale} & 29.784 & 1.051e-3 & 27.876 & 1.631e-3 & 28.177 & 1.522e-3 & 29.043 & 1.247e-3 & 29.264 & 1.185e-3 \\ \cmidrule(lr){1-11}
\textbf{Fourier (Ours)} & \textbf{36.091} & \textbf{2.460e-4} & \textbf{33.773} & \textbf{4.195e-4} & \textbf{32.811} & \textbf{5.235e-4} & \textbf{35.589} & \textbf{2.761e-4} & \textbf{36.392} & \textbf{2.295e-4}  \\ 
\specialrule{1pt}{2pt}{2pt}
\end{tabular}
\label{tbl:experiment results}
\end{table*}

\subsection{Experimental Settings}
\paragraph{Experimental Setups} We implement our ImpliSat framework and other baselines using \textsc{PyTorch} 2.3.1~\cite{paszke2019pytorch} in a single \textsc{NVIDIA RTX 3090 GPU}. We set $L=6$, $n=256$, and $m=32$. The hypernetwork $\Pi$ is also an MLP, with 3 layers and 64 neurons. All models are trained for 10,000 iterations with approximately $200K$ trainable parameters (1MB per model checkpoint, around $10\times$ smaller than the original image). We use the Adam optimizer~\cite{kingma2014adam} and early stopping.

\paragraph{Datasets}
We compile a dataset of five multispectral Sentinel-2~\cite{fletcher2012sentinel} images, each capturing diverse environments: London (UK), Seoul (South Korea), Merapi (Indonesia), Hawaii (USA), Cairo (Egypt).  Each MSI has a size of 9.4MB. The Table~\ref{tbl:dataset_coord} provides the geographical coordinates (longitude and latitude) of each location. Those locations are chosen to evaluate the ability of ImpliSat framework to compress MSI with various complexity and difficulty.

\subsection{Comparison with Existing Modulated INRs}
We compare our Fourier modulation approach with existing modulation techniques, specifically shift modulation used in~\cite{dupont2022coin++, dupont2022data, bauer2023spatial} and scale modulation proposed in~\cite{dupont2022coin++}. Shift modulation involves adding a learnable bias term $\mu$ to the output of each MLP layer ($h_{l+1} = \text{sin}(W_l \cdot h_l + b_l + {\mu}_l)$)
Meanwhile,  scale modulation extends the shift modulation by scaling the output of each layer using modulation vector $\kappa$ ($h_{l+1} = \text{sin}(\kappa_l \odot (W_l \cdot h_l + b_l))$).

All approaches are evaluated using PSNR and MSE to measure the accuracy of reconstructed images against the ground truth. 
As shown in Table~\ref{tbl:experiment results}, Fourier modulation consistently outperforms both shift and scale modulation. Notably, in complex areas such as Seoul, our approach demonstrates a 20\% performance improvement over the baseline models.
\begin{figure}[t]
\centering
    \scriptsize
    \includegraphics[width=0.34\textwidth]{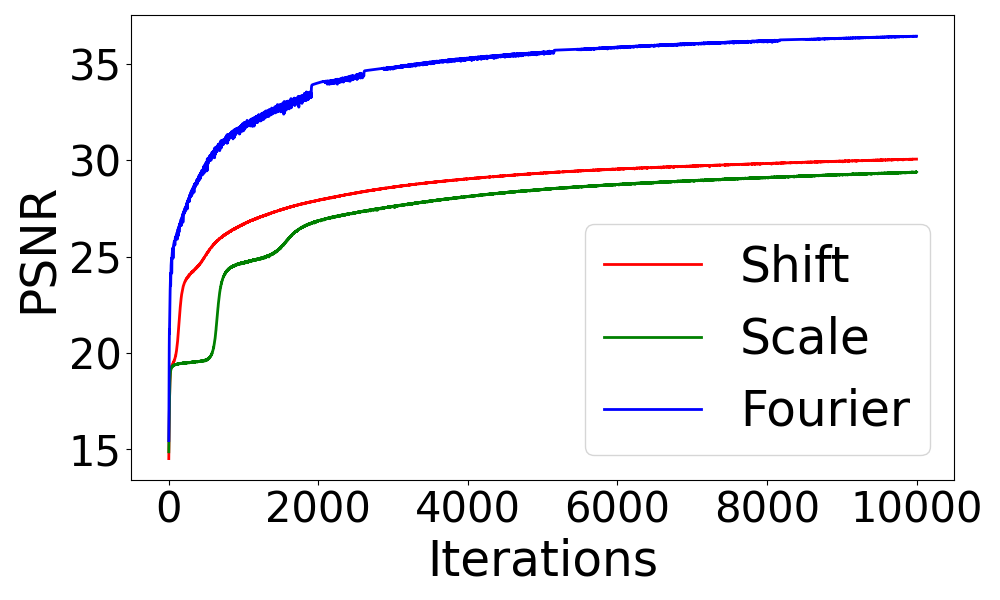}
    \caption{Comparison of PSNR across iterations for Shift, Scale, and Fourier (ours) methods on Cairo.}
    \label{fig:compare_20}
\end{figure}
Fig.~\ref{fig:main_results} presents a visualization comparison for each MSI\footnote{Due to space constraints, we only show one band for each MSI in Fig.~\ref{fig:main_results}}, where the rows (top to bottom) correspond to Cairo, Merapi, London, Seoul, and Hawaii. 
The first and second rows show bands with 10m GSD, the third and fourth rows show bands with 20m GSD, and the last one is a band with 60m GSD. These results further illustrate that our Fourier modulation captures sharper and more detailed structures in complex areas, such as the bridges and buildings visible in the Seoul B8A and segmentation between land patches in London B7. Unlike shift and scale modulation, which tend to blur fine details, our approach preserves clear edges and sharp features, leading to significantly better reconstructions regardless of the GSD. 
In Fig.~\ref{fig:compare_20}, we show the PSNR comparison between Fourier modulation and the baseline modulations throughout the training process on the Cairo dataset. Fourier modulation has much better convergence rate, as evidenced by the rapid increase in PSNR, and continues to maintain superior performance as training progresses. 



\subsection{Analysis on Fourier Modulations}
\begin{figure}[t]
\subfigure[10m GSD (B2)]{\includegraphics[width=0.31\columnwidth]{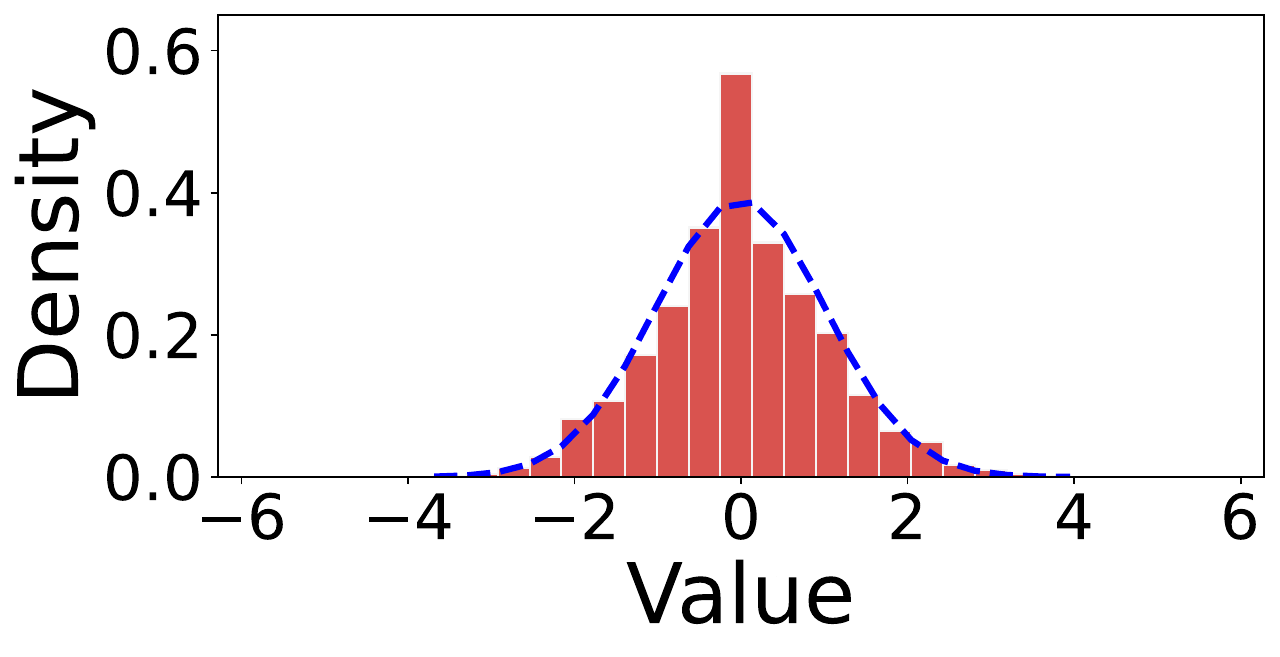}}\hfill
\subfigure[20m GSD (B7)]{\includegraphics[width=0.31\columnwidth]{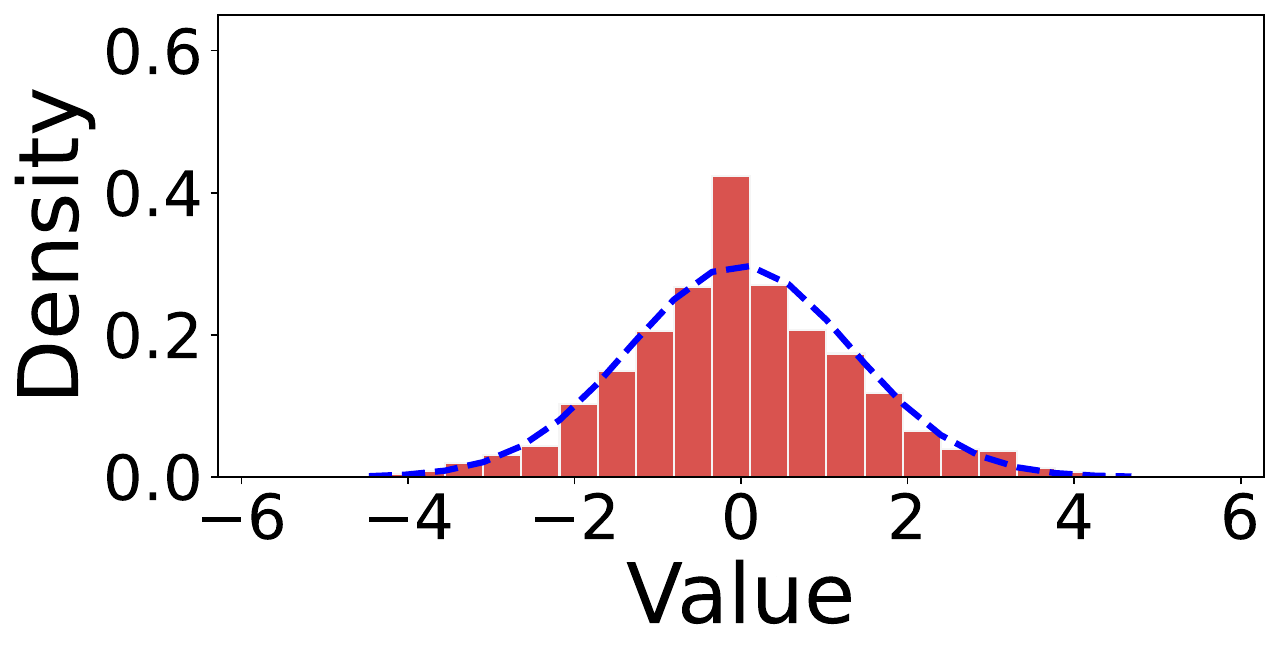}}\hfill
\subfigure[60m GSD (B10)]{\includegraphics[width=0.31\columnwidth]{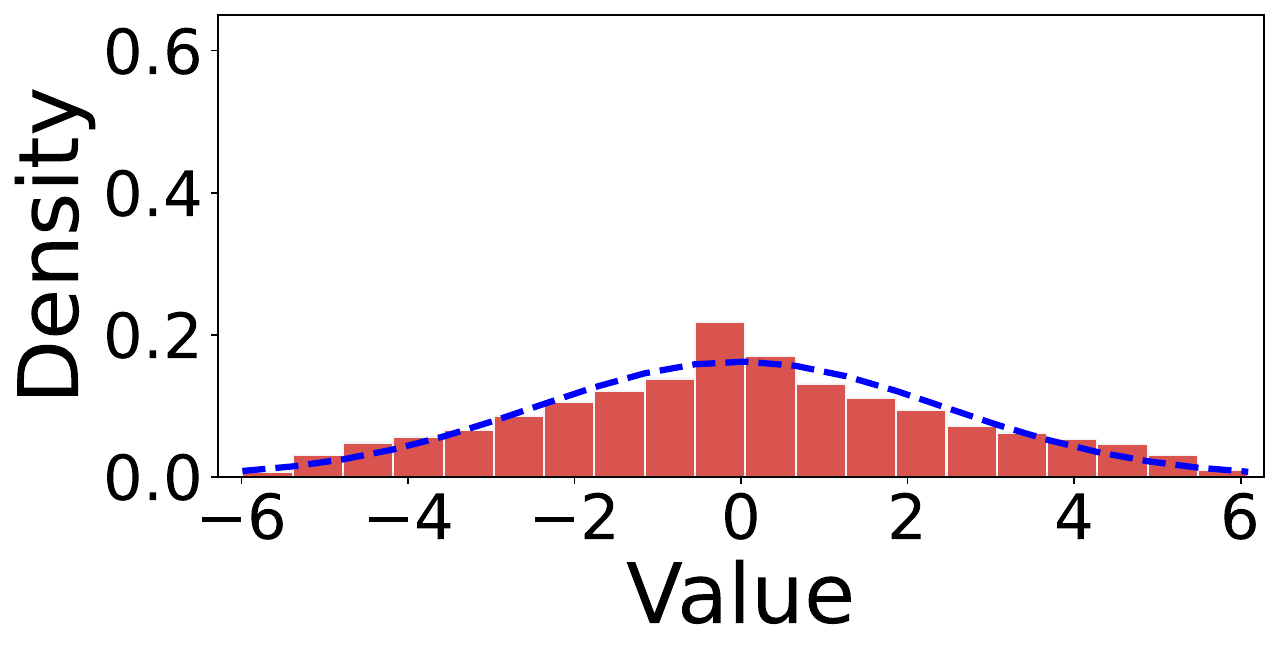}}\\
\caption{Density distribution of the frequency component of Fourier modulations generated by the hypernetwork for different spatial resolutions.}
\label{fig:dist_fourier}
\end{figure}

Fig.~\ref{fig:dist_fourier} presents histograms depicting the density distribution of the frequency components ($\Omega \odot \mathcal{Z}$) for 10m, 20m, and 60m GSD. The distribution of Fourier modulation frequencies varies notably with different resolutions.
For 10m GSD (cf. Fig.~\ref{fig:dist_fourier}(a)), the values are more concentrated near zero, indicating that precise modulation has been applied to effectively represent high-resolution bands.
In contrast, the 60m GSD (cf. Fig.~\ref{fig:dist_fourier}(c)) distribution is broader, suggesting that a more variable modulation strategy has been adopted to accommodate the characteristics of low-resolution bands.
These results suggest that the hypernetwork successfully adjusts the frequency content of the Fourier modulations according to the resolution-specific needs of the data. This adaptive capability highlights the flexibility of the Fourier modulation approach, enabling ImpliSat to address the diverse challenges of multispectral satellite imagery.

\section{Conclusion}
In this paper, we present ImpliSat, a framework that uses Fourier-modulated INR to compress and reconstruct multispectral satellite images. By dynamically adjusting Fourier bases to match band-specific resolutions and spectral properties, ImpliSat outperforms traditional modulation methods, particularly in high-resolution and complex environments. The experimental results demonstrate the effectiveness of this approach for efficient MSI data compression while maintaining the original quality.

\small
\bibliographystyle{IEEEtranN}
\bibliography{references}

\begin{thebibliography}{25}
\providecommand{\natexlab}[1]{#1}
\providecommand{\url}[1]{#1}
\csname url@samestyle\endcsname
\providecommand{\newblock}{\relax}
\providecommand{\bibinfo}[2]{#2}
\providecommand{\BIBentrySTDinterwordspacing}{\spaceskip=0pt\relax}
\providecommand{\BIBentryALTinterwordstretchfactor}{4}
\providecommand{\BIBentryALTinterwordspacing}{\spaceskip=\fontdimen2\font plus
\BIBentryALTinterwordstretchfactor\fontdimen3\font minus \fontdimen4\font\relax}
\providecommand{\BIBforeignlanguage}[2]{{%
\expandafter\ifx\csname l@#1\endcsname\relax
\typeout{** WARNING: IEEEtranN.bst: No hyphenation pattern has been}%
\typeout{** loaded for the language `#1'. Using the pattern for}%
\typeout{** the default language instead.}%
\else
\language=\csname l@#1\endcsname
\fi
#2}}
\providecommand{\BIBdecl}{\relax}
\BIBdecl

\bibitem[Yang et~al.(2013)Yang, Gong, Fu, Zhang, Chen, Liang, Xu, Shi, and Dickinson]{yang2013role}
J.~Yang, P.~Gong, R.~Fu, M.~Zhang, J.~Chen, S.~Liang, B.~Xu, J.~Shi, and R.~Dickinson, ``The role of satellite remote sensing in climate change studies,'' \emph{Nature climate change}, vol.~3, no.~10, pp. 875--883, 2013.

\bibitem[Norris et~al.(2016)Norris, Allen, Evan, Zelinka, O’Dell, and Klein]{norris2016evidence}
J.~R. Norris, R.~J. Allen, A.~T. Evan, M.~D. Zelinka, C.~W. O’Dell, and S.~A. Klein, ``Evidence for climate change in the satellite cloud record,'' \emph{Nature}, vol. 536, no. 7614, pp. 72--75, 2016.

\bibitem[Masek(2001)]{masek2001stability}
J.~G. Masek, ``Stability of boreal forest stands during recent climate change: evidence from landsat satellite imagery,'' \emph{Journal of biogeography}, vol.~28, no.~8, pp. 967--976, 2001.

\bibitem[Hassan-Esfahani et~al.(2015)Hassan-Esfahani, Torres-Rua, Jensen, and McKee]{hassan2015assessment}
L.~Hassan-Esfahani, A.~Torres-Rua, A.~Jensen, and M.~McKee, ``Assessment of surface soil moisture using high-resolution multi-spectral imagery and artificial neural networks,'' \emph{Remote Sensing}, vol.~7, no.~3, pp. 2627--2646, 2015.

\bibitem[Dwyer et~al.(2000)Dwyer, Pinnock, Gr{\'e}goire, and Pereira]{dwyer2000global}
E.~Dwyer, S.~Pinnock, J.-M. Gr{\'e}goire, and J.~Pereira, ``Global spatial and temporal distribution of vegetation fire as determined from satellite observations,'' \emph{International Journal of Remote Sensing}, vol.~21, no. 6-7, pp. 1289--1302, 2000.

\bibitem[Ma et~al.(2008)Ma, Duan, Gu, and Zhang]{ma2008detecting}
R.~Ma, H.~Duan, X.~Gu, and S.~Zhang, ``Detecting aquatic vegetation changes in taihu lake, china using multi-temporal satellite imagery,'' \emph{Sensors}, vol.~8, no.~6, pp. 3988--4005, 2008.

\bibitem[Pettorelli(2019)]{pettorelli2019satellite}
N.~Pettorelli, \emph{Satellite remote sensing and the management of natural resources}.\hskip 1em plus 0.5em minus 0.4em\relax Oxford University Press, 2019.

\bibitem[Sitzmann et~al.(2020)Sitzmann, Martel, Bergman, Lindell, and Wetzstein]{sitzmann2020implicit}
V.~Sitzmann, J.~Martel, A.~Bergman, D.~Lindell, and G.~Wetzstein, ``Implicit neural representations with periodic activation functions,'' \emph{Advances in neural information processing systems}, vol.~33, pp. 7462--7473, 2020.

\bibitem[Strümpler et~al.(2022)Strümpler, Postels, Yang, Gool, and Tombari]{strumpler2022inrcompress}
Y.~Strümpler, J.~Postels, R.~Yang, L.~V. Gool, and F.~Tombari, ``Implicit neural representations for image compression,'' in \emph{ECCV}, 2022.

\bibitem[Mudiyanselage et~al.(2025)Mudiyanselage, Cho, Jo, Park, and Lee]{mudiyanselage2025unveiling}
U.~B. Mudiyanselage, W.~Cho, M.~Jo, N.~Park, and K.~Lee, ``Unveiling the potential of superexpressive networks in implicit neural representations,'' \emph{arXiv preprint arXiv:2503.21166}, 2025.

\bibitem[Mildenhall et~al.(2020)Mildenhall, Srinivasan, Tancik, Barron, Ramamoorthi, and Ng]{mildenhall2020nerf}
B.~Mildenhall, P.~P. Srinivasan, M.~Tancik, J.~T. Barron, R.~Ramamoorthi, and R.~Ng, ``Nerf: Representing scenes as neural radiance fields for view synthesis,'' in \emph{ECCV}, 2020.

\bibitem[Pumarola et~al.(2021)Pumarola, Corona, Pons-Moll, and Moreno-Noguer]{pumarola2021d}
A.~Pumarola, E.~Corona, G.~Pons-Moll, and F.~Moreno-Noguer, ``D-nerf: Neural radiance fields for dynamic scenes,'' in \emph{Proceedings of the IEEE/CVF Conference on Computer Vision and Pattern Recognition}, 2021, pp. 10\,318--10\,327.

\bibitem[Cho et~al.(2025{\natexlab{a}})Cho, Jo, Lee, and Park]{cho2025neural}
W.~Cho, M.~Jo, K.~Lee, and N.~Park, ``Neural functions for learning periodic signal,'' in \emph{The Thirteenth International Conference on Learning Representations}, 2025.

\bibitem[Wallace(1992)]{wallace1992jpeg}
G.~K. Wallace, ``The jpeg still picture compression standard,'' \emph{Communications of the ACM}, vol.~34, no.~4, pp. 30--44, 1992.

\bibitem[Boutell(1997)]{boutell1997png}
T.~Boutell, \emph{PNG: The Definitive Guide}.\hskip 1em plus 0.5em minus 0.4em\relax O'Reilly Media, Inc., 1997.

\bibitem[Dupont et~al.(2022{\natexlab{a}})Dupont, Loya, Alizadeh, Goli'nski, Teh, and Doucet]{dupont2022coin++}
E.~Dupont, H.~Loya, M.~Alizadeh, A.~Goli'nski, Y.~W. Teh, and A.~Doucet, ``Coin++: Neural compression across modalities,'' \emph{TMLR}, 2022.

\bibitem[Hu et~al.(2022)Hu, Shen, Wallis, Allen-Zhu, Li, Wang, Wang, and Chen]{hu2021lora}
E.~J. Hu, Y.~Shen, P.~Wallis, Z.~Allen-Zhu, Y.~Li, S.~Wang, L.~Wang, and W.~Chen, ``Lo{RA}: Low-rank adaptation of large language models,'' in \emph{International Conference on Learning Representations}, 2022.

\bibitem[Cho et~al.(2024)Cho, Lee, Rim, and Park]{cho2024hypernetwork}
W.~Cho, K.~Lee, D.~Rim, and N.~Park, ``Hypernetwork-based meta-learning for low-rank physics-informed neural networks,'' \emph{Advances in Neural Information Processing Systems}, vol.~36, 2024.

\bibitem[Cho et~al.(2025{\natexlab{b}})Cho, Lee, Park, Rim, and Welper]{cho2025fastlrnr}
W.~Cho, K.~Lee, N.~Park, D.~Rim, and G.~Welper, ``Fastlrnr and sparse physics informed backpropagation,'' \emph{Results in Applied Mathematics}, vol.~25, p. 100547, 2025.

\bibitem[Shi et~al.(2024)Shi, Zhou, and Gu]{shi2024improved}
K.~Shi, X.~Zhou, and S.~Gu, ``Improved implicit neural representation with fourier reparameterized training,'' in \emph{Proceedings of the IEEE/CVF Conference on Computer Vision and Pattern Recognition}, 2024, pp. 25\,985--25\,994.

\bibitem[Paszke et~al.(2019)Paszke, Gross, Massa, Lerer, Bradbury, Chanan, Killeen, Lin, Gimelshein, Antiga, et~al.]{paszke2019pytorch}
A.~Paszke, S.~Gross, F.~Massa, A.~Lerer, J.~Bradbury, G.~Chanan, T.~Killeen, Z.~Lin, N.~Gimelshein, L.~Antiga \emph{et~al.}, ``Pytorch: An imperative style, high-performance deep learning library,'' \emph{Advances in neural information processing systems}, vol.~32, 2019.

\bibitem[Kingma and Ba(2014)]{kingma2014adam}
D.~P. Kingma and J.~Ba, ``Adam: A method for stochastic optimization,'' \emph{CoRR}, 2014.

\bibitem[Fletcher(2012)]{fletcher2012sentinel}
K.~Fletcher, \emph{SENTINEL 2: ESA's Optical High-Resolution Mission for GMES Operational Services}.\hskip 1em plus 0.5em minus 0.4em\relax European Space Agency, 2012.

\bibitem[Dupont et~al.(2022{\natexlab{b}})Dupont, Kim, Eslami, Rezende, and Rosenbaum]{dupont2022data}
E.~Dupont, H.~Kim, S.~Eslami, D.~Rezende, and D.~Rosenbaum, ``From data to functa: Your data point is a function and you can treat it like one,'' in \emph{ICML}, 2022.

\bibitem[Bauer et~al.(2023)Bauer, Dupont, Brock, Rosenbaum, Schwarz, and Kim]{bauer2023spatial}
M.~Bauer, E.~Dupont, A.~Brock, D.~Rosenbaum, J.~R. Schwarz, and H.~Kim, ``Spatial functa: Scaling functa to imagenet classification and generation,'' \emph{arXiv preprint arXiv:2302.03130}, 2023.

\end{thebibliography}

\end{document}